\documentclass{article}
\pdfoutput=1
\usepackage{spconf,amsmath,graphicx}
\usepackage{multirow}
\usepackage{tabularx}
\usepackage{hyperref}

\title{How Deep Neural Networks Can Improve \\Emotion Recognition on Video Data}
\name{Pooya Khorrami$^{1}$\thanks{The MIT Lincoln Laboratory part of this collaborative work is sponsored by the Assistant Secretary of Defense for Research \& Engineering under Air Force Contract FA8721-05-C-0002. The Tesla K40 GPU used for this research was donated by the NVIDIA Corporation.}, Tom Le Paine$^{1}$, Kevin Brady$^{2}$, Charlie Dagli$^{2}$, Thomas S. Huang$^{1}$}
 \address{$^{1}$ Beckman Institute, University of Illinois at Urbana-Champaign\\
       $^{2}$ MIT Lincoln Laboratory\\
       $^{1}$ \texttt{\{pkhorra2, paine1, t-huang1\}@illinois.edu} \\ $^{2}$ \texttt{\{kbrady, dagli\}@ll.mit.edu}}

\begin{document}
%
\maketitle
\begin{abstract}
We consider the task of dimensional emotion recognition on video data using deep learning. While several previous methods have shown the benefits of training temporal neural network models such as recurrent neural networks (RNNs) on hand-crafted features, few works have considered combining convolutional neural networks (CNNs) with RNNs. In this work, we present a system that performs emotion recognition on video data using both CNNs and RNNs, and we also analyze how much each neural network component contributes to the system's overall performance. We present our findings on videos from the Audio/Visual+Emotion Challenge (AV+EC2015). In our experiments, we analyze the effects of several hyperparameters on overall performance while also achieving superior performance to the baseline and other competing methods.
\end{abstract}

\section{Introduction}
\label{sec:intro}

For several decades, emotion recognition has remained one of the of the most important problems in the field of human computer interaction. A large portion of the community has focused on \textit{categorical models} which try to group emotions into discrete categories. The most famous categories are the six basic emotions originally proposed by Ekman in \cite{ekman1987universals, ekman1994strong}: anger, disgust, fear, happiness, sadness, and surprise. These emotions were selected because they were all perceived similarly regardless of culture. 

Several datasets have been constructed to evaluate automatic emotion recognition systems such as the extended Cohn-Kanade (CK+) dataset \cite{lucey2010extended} the MMI facial expression database \cite{valstar2010induced} and the Toronto Face Dataset (TFD) \cite{susskind2010toronto}. In the last few years, several methods based on hand-crafted and, later, learned features \cite{shan2009facial, liu2013aware, liu2014facial, khorrami2015deep} have performed quite well in recognizing the six basic emotions. Unfortunately, these six basic emotions do not cover the full range of emotions that a person can express.

An alternative way to model the space of possible emotions is to use a \textit{dimensional} approach \cite{russell1977evidence} where a person's emotions can be described using a low-dimensional signal (typically 2 or 3 dimensions). The most common dimensions are (i) arousal and (ii) valence. Arousal measures how engaged or apathetic a subject appears while valence measures how positive or negative a subject appears. 

Dimensional approaches have two advantages over categorical approaches. The first being that dimensional approaches can describe a larger set of emotions. Specifically, the arousal and valence scores define a two dimensional plane while the six basic emotions are represented as points in said plane. The second advantage is dimensional approaches can output time-continuous labels which allows for more realistic modeling of emotion over time. This could be particularly useful for representing video data. 
	
Given the success of deep neural networks on datasets with categorical labels \cite{kahou2013combining,  khorrami2015deep}, one can ask the very natural question: is it possible to train a neural network to learn a representation that is useful for dimensional emotion recognition in video data? 


In this paper, we will present two different frameworks for training an emotion recognition system using deep neural networks. The first is a single frame convolutional neural network (CNN) and the second is a combination of CNN and a recurrent neural network (RNN) where each input to the RNN is the fully-connected features of a single frame CNN. While many works \cite{ringeval2014prediction, ringeval2015av+, chen2015multi, he2015multimodal} have trained recurrent neural networks (RNNs) on hand-crafted features, few works \cite{chao2015long, ebrahimi2015recurrent} have considered the effects of using a CNN as input to the RNN. Our work, developed concurrently with \cite{chao2015long, ebrahimi2015recurrent}, demonstrates  the benefits of using a CNN+RNN model and also shows how much the CNN and the RNN individually contribute to the overall performance.

Thus, the main contributions of this work are as follows:
\begin{enumerate}
\item We train both a single-frame CNN and a CNN+RNN model and analyze their effectiveness on the dimensional emotion recognition task. We also conduct extensive analysis on the various hyperparameters of the CNN+RNN model to support our chosen architecture.
\item We evaluate our models on the AV+EC 2015 dataset \cite{ringeval2015av+} and demonstrate that both of our techniques can achieve comparable or superior performance to the baseline model and other competing methods.
\end{enumerate}

\section{Dataset}
\label{sec:dataset}
The AV+EC 2015 \cite{ringeval2015av+} corpus uses data from the RECOLA dataset \cite{ringeval2013introducing}, a multimodal corpus designed to monitor subjects as they worked
in pairs remotely to complete a collaborative task. The type of modalities include: audio, video, 
electro-cardiogram (ECG) and electro-dermal activity (EDA). These signals were recorded for 27 French-speaking subjects. The dataset contains two types of dimensional labels (arousal and valence) which were annotated by 6 people. Each dimensional label ranges from $ \left[-1, 1\right] $. The dataset is partitioned into three sets: train, development, and test, each containing 9 subjects. 

In our experiments, we focus on predicting the valence score using just the video modality. Also, since the test set labels were not readily available, we evaluate all of our experiments on the development set. We evaluate our techniques by computing three metrics: (i) Root Mean Square Error (RMSE) (ii) Pearson Correlation Coefficient (CC) and (iii) Concordance Correlation Coefficient (CCC). The Concordance Correlation Coefficient tries to measure the agreement between two variables using the following expression: 
\begin{equation}
\rho_c = \frac{2 \rho \sigma_x \sigma_y}{\sigma_x^2 + \sigma_y^2 + (\mu_x - \mu_y)^2}
\label{eq:ccc}
\end{equation}

where $\rho$ is the Pearson correlation coefficient, $\sigma_x^2$ and $\sigma_y^2$ are the variance of the predicted and ground truth values respectively and $\mu_x$ and $\mu_y$ are their means, respectively.
The strongest method is selected based on whichever obtains the highest CCC value.

\section{Our Approach}
\label{sec:method}
\begin{figure}[t!]
\begin{minipage}[b]{\linewidth}
  \centering
 \centerline{\includegraphics[trim = 0mm 5mm 0mm 0mm, clip, width=1.0\textwidth, keepaspectratio]{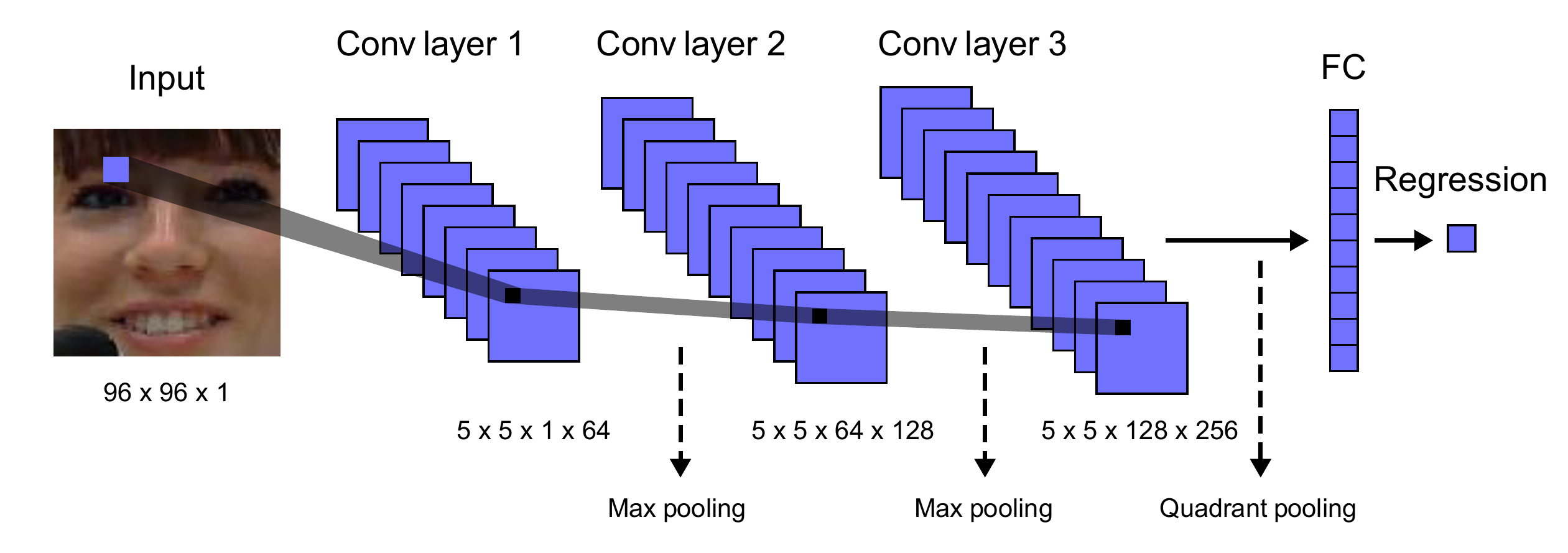}}
\end{minipage}
\caption{Single Frame CNN Architecture - Similar to the network in \cite{khorrami2015deep}, our network consists of three convolutional layers containing 64, 128, and 256 filters, respectively, each of size 5x5 followed by ReLU (Rectified Linear Unit) activation functions. We add 2x2 max pooling layers after the first two convolutional layers and quadrant pooling after the third. The three convolutional layers are followed by a fully-connected layer containing 300 hidden units and a linear regression layer which approximates the valence score.}
\label{fig:cnn_network_architecture}
\end{figure} 

\begin{figure}[t!]
\begin{minipage}[b]{\linewidth}
  \centering
 \centerline{\includegraphics[trim = 0mm 5mm 0mm 0mm, clip, width=1.0\textwidth, keepaspectratio]{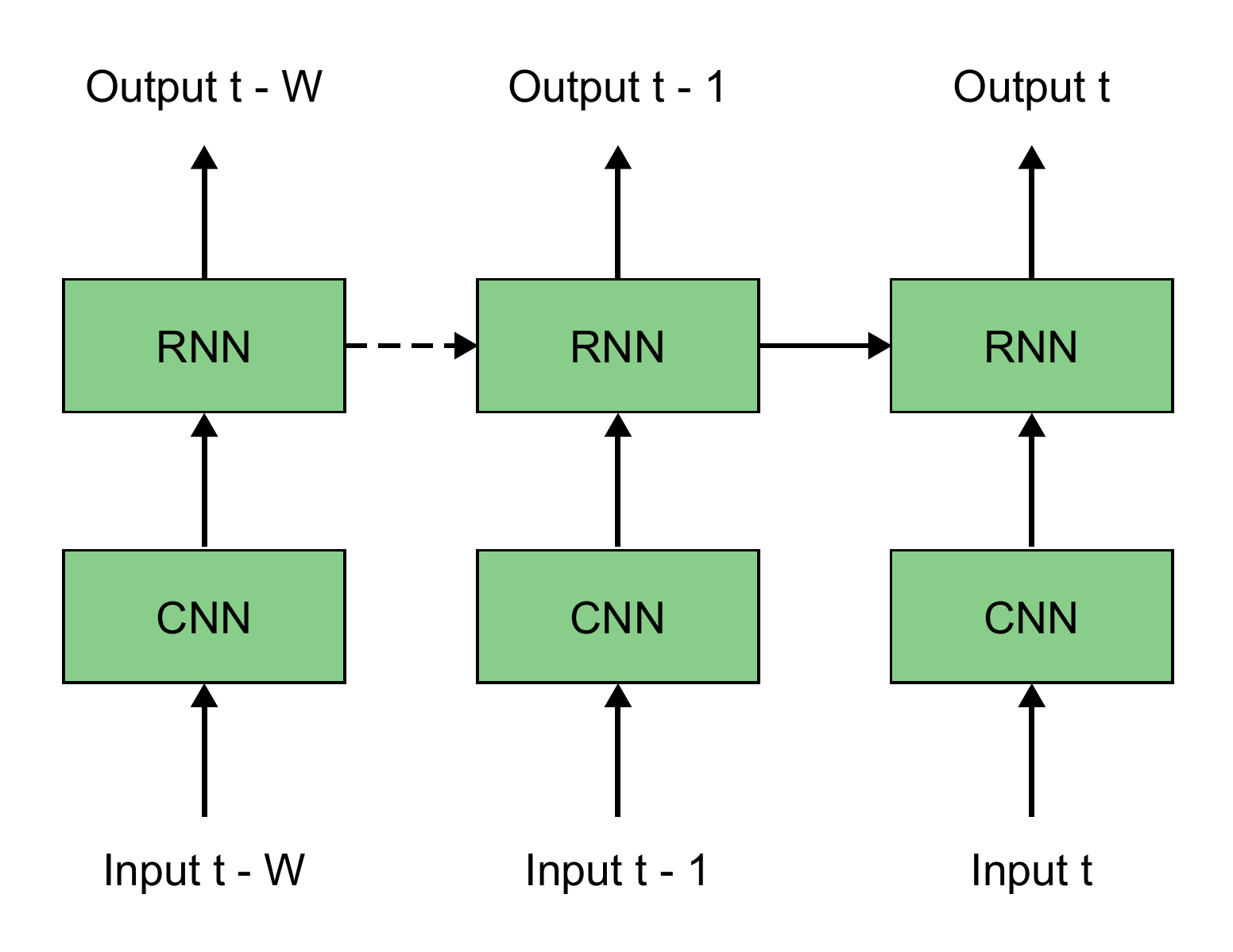}}  
\end{minipage}
\caption{CNN+RNN Network Architecture - Given a time $t$ in a video, we extract a window of length W frames $(\left[t-W, t\right])$. We model our single frame CNN as a feature extractor by fixing all of the parameters and removing the top regression layer. We then pass each frame within the window to the CNN and extract a 300 dimensional feature for every frame, each of which is passed as an input to one node of the RNN. We then take the valence score generated by the RNN at time $t$.}
\label{fig:cnn_rnn_network_architecture}
\end{figure} 

\subsection{Single Frame Regression CNN}
The first model that we train is a single frame CNN. At each time point in a video, we pass the corresponding video frame through a CNN, shown visually in Figure \ref{fig:cnn_network_architecture}. The CNN has 3 convolutional layers consisting of 64, 128, and 256 filters respectively, each of size 5x5. The first two convolutional layers are followed by 2x2 max pooling while the third layer is followed by quadrant pooling.  After the convolutional layers is a fully-connected layer with 300 hidden units and a linear regression layer to estimate the valence label. We use the mean squared error (MSE) as our cost function. 

All of the CNNs were trained using stochastic gradient descent with batch size of 128, momentum equal to 0.9, and weight decay of 1e-5. We used a constant learning of 0.01 and did not use any form of annealing. All of our CNN models were trained using the anna software library \footnote{\url{https://github.com/ifp-uiuc/anna}}. 

\subsection{Adding Recurrent Neural Networks (RNNs)}
Despite having the ability to learn useful features directly from the video data, the single frame regression CNN completely ignores temporal information. Similar to the model in \cite{ebrahimi2015recurrent}, we propose to incorporate the temporal information by using a recurrent neural network (RNN) to propagate information from one time point to next. 

We first model the CNN as a feature extractor by fixing all of its parameters and removing the regression layer. Now, when a frame is passed to the CNN, we extract a 300 dimensional vector from the fully-connected layer. For a given time t, we take $W$ frames from the past (i.e. $\left[t-W, t\right]$). We then pass each frame from time $t-W$ to $t$ to the CNN and extract W vectors in total, each length of 300. Each of these vectors is then passed as input to a node of the RNN. Each node in the RNN then regresses the output valence label. We visualize the model in Figure \ref{fig:cnn_rnn_network_architecture}. Once again we use the mean squared error (MSE) as our cost function during optimization.

We train our RNN models with stochastic gradient descent with a constant learning rate of 0.01, a batch size of 128 and momentum equal to 0.9. All of the RNNs in our experiments were trained using the Lasagne library \footnote{\url{https://github.com/Lasagne/Lasagne}}.

\section{Experiments}
\label{sec:experiments}
\subsection{Data Preprocessing}
When preparing the video data, we first detect the face in each video frame using face and landmark detector in Dlib-ml \cite{king2009dlib}. Frames where the face detector missed were dropped and their valence scores were later computed by linearly interpolating the scores from adjacent frames. We then map the detected landmark points to pre-defined pixel locations in order to ensure correspondence between frames. After normalizing the eye and nose coordinates, we apply mean subtraction and contrast normalization prior to passing each face image through the CNN.

\subsection{Single Frame CNN vs. CNN+RNN}
Table \ref{tab:our_models} shows how well our single frame regression CNN and our CNN+RNN architecture perform at predicting valence scores of subjects in the development set of the AV+EC 2015 dataset. When training our single frame CNN, we consider two forms of regularization: dropout (D) with probability 0.5 and data augmentation (A) in form of flips and color changes. For our CNN+RNN model, we use a single layer RNN with 100 units in the hidden layer and a temporal window of size 100 frames. We consider two types of nonlinearities: hyperbolic tangent (tanh) and rectified linear unit (ReLU). 

From Table \ref{tab:our_models}, we can see, not surprisingly, that adding regularization improves the performance of the CNN. Most notably, we see that our CNN model with dropout (CNN+D) outperforms the baseline LSTM model trained on LBP-TOP features \cite{ringeval2015av+} (CCC = 0.326 vs. 0.273). Finally, when incorporating temporal information using the CNN+RNN model, we can achieve a significant performance gain over the single frame CNN. 

\begin{figure}[t!]
\begin{minipage}[b]{\linewidth}
  \centering
  \centerline{\includegraphics[trim = 5mm 0mm 5mm 5mm, clip, width=1.0\textwidth, keepaspectratio]{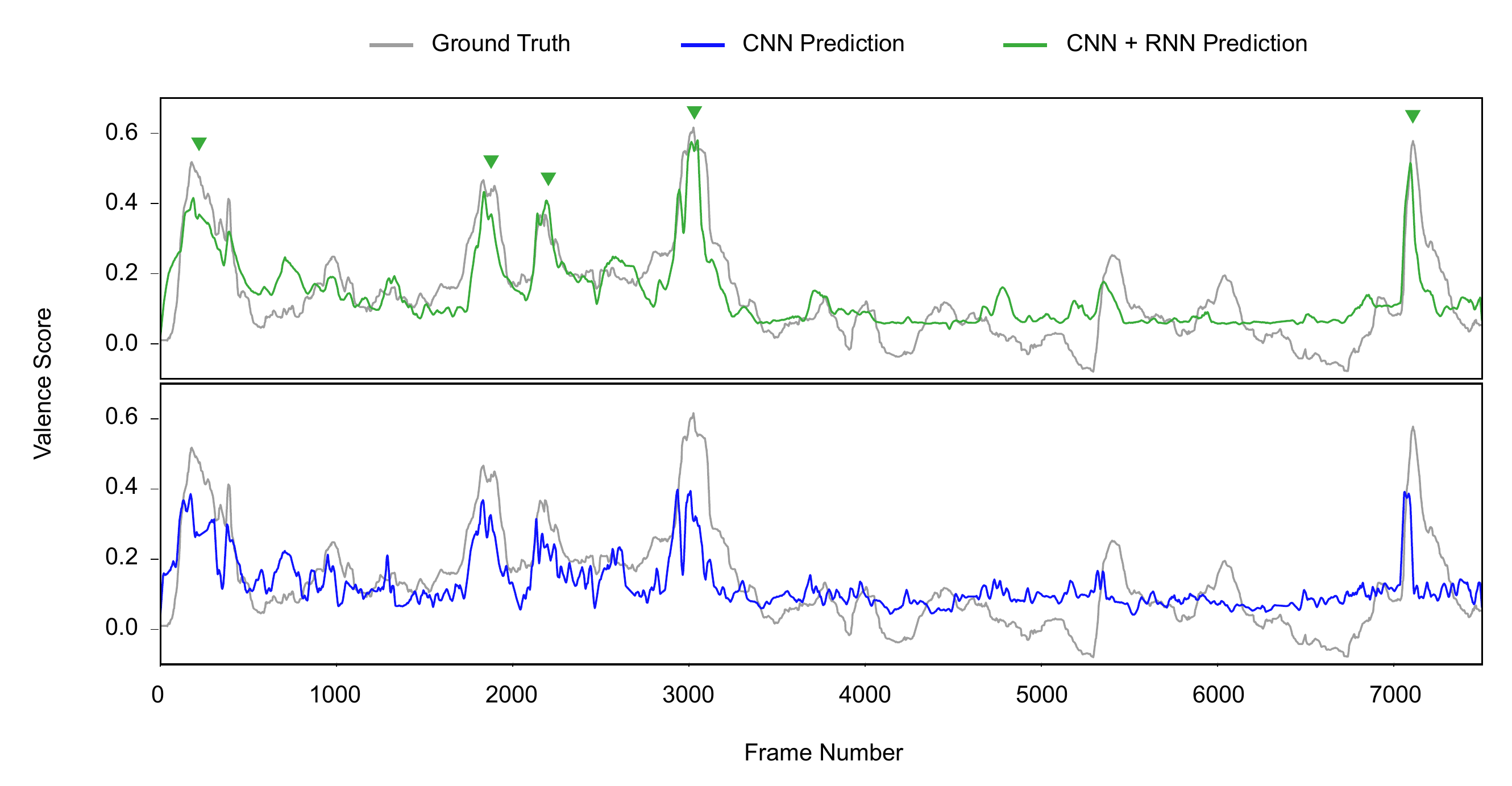}}  
\end{minipage}
\caption{Valence score predictions of the single frame CNN and the CNN+RNN model for one subject in the AV+EC 2015 development set - Notice that the CNN+RNN model appears to smooth the scores outputted by the single frame CNN and seems to approximate the ground truth more accurately, specifically the peaks (arrows). (Best viewed in color).}
\label{fig:valence_pred_signal}
\end{figure}

In Figure \ref{fig:valence_pred_signal}, we plot the valence scores predicted by both our single frame CNN and the CNN+RNN model for one of the videos in the development set. From this chart, we can clearly see the advantages of using temporal information. The CNN+RNN model appears to model the ground truth more accurately and generate a smoother prediction than the single frame regression CNN. 

\begin{table}[t!]
\caption{Performance comparison between: (i) Baseline method with hand-crafted features (ii) Single frame CNN with different levels of regularization (iii) Single frame CNN with an RNN connecting each time point (A = Data Augmentation, D = Dropout) }
\begin{center}
    \begin{tabular}{ | l | c | c | c |}
    \hline
    \bf Method & \bf RMSE & \bf CC & \bf CCC \\ \hline
    Baseline \cite{ringeval2015av+} & 0.117 & 0.358 & 0.273 \\ \hline
    \hline
    CNN  & 0.121 & 0.341 & 0.242 \\ 
    CNN+D & 0.113 & 0.426 & 0.326 \\ 
    CNN+A & 0.125 & 0.349 & 0.270 \\ 
    CNN+AD & 0.118 & 0.405 & 0.309 \\ \hline 
    \hline
    CNN+RNN - tanh &  0.111 & 0.518 & 0.492 \\ 
    \bf CNN+RNN - ReLU & \bf 0.108 & \bf 0.544 & \bf 0.506 \\ \hline
    \end{tabular}
\end{center}
\label{tab:our_models}
\end{table}

\subsection{Hyperparameter Analysis}
We study the effects of several hyperparameters in the CNN+RNN model, namely the number of hidden units, the length of the temporal window, and the number of hidden layers in the RNN. The results are shown in Tables \ref{tab:num_hidden_units}, \ref{tab:window_length}, and \ref{tab:num_hidden_layers} respectively. Based on our results in Table \ref{tab:num_hidden_units}, we conclude that it is best to have $\approx$100 hidden units given that both $h=50$ and $h=200$ resulted in decreases in performance. Similarly, for the temporal window length, we see that a window of length 100 frames $(\approx 4 \text{ seconds})$ appears to yield the highest CCC score, while reducing the window to 25 frames (1 second) and increasing it to 150 frames (6 seconds) both lead to significant decreases in performance. In Table \ref{tab:num_hidden_layers}, we see that increasing the number of hidden layers yields a small improvement in performance. Thus, based on our experiments, our best performing model had 3 hidden layers with a window length of W=100 frames, 100 hidden units in the first two recurrent layers and 50 in the third, and a ReLU as its nonlinearity.


\begin{table}[t!]
\caption{Effect of Changing Number of Hidden Units}
\begin{center}
    \begin{tabular}{ | l | c | c | c |}
    \hline
    \bf Method & \bf RMSE & \bf CC & \bf CCC \\ \hline
    CNN+RNN - h=50 & 0.110 & 0.519 & 0.485 \\ 
    \bf CNN+RNN - h=100 & \bf 0.108 & \bf 0.544 & \bf 0.506 \\ 
    CNN+RNN - h=150 & 0.112 & 0.529 & 0.494 \\ 
    CNN+RNN - h=200 & 0.108 & 0.534 & 0.495 \\ \hline
    \end{tabular}
\end{center}
\label{tab:num_hidden_units}
\end{table}


\begin{table}[t!]
\caption{Effect of Changing Temporal Window Length (i.e. number of frames used by the RNN)}
\begin{center}
    \begin{tabular}{ | l | c | c | c |}
    \hline
    \bf Method & \bf RMSE & \bf CC & \bf CCC \\ \hline
    CNN+RNN - W=25 & 0.111 & 0.501 & 0.474 \\ 
    CNN+RNN - W=50 & 0.112 & 0.526 & 0.492 \\ 
    CNN+RNN - W=75 & 0.111 & 0.528 & 0.498 \\ 
    \bf CNN+RNN - W=100 & \bf 0.108 & \bf 0.544 & \bf 0.506 \\ 
    CNN+RNN - W=150 & 0.110 & 0.521 & 0.485 \\ \hline
    \end{tabular}
\end{center}
\label{tab:window_length}
\end{table}


\begin{table}[t!]
\caption{Effect of Changing the Number of Hidden Layers in the RNN}
\begin{center}
\resizebox{0.95\columnwidth}{!}{
    \begin{tabular}{ | l | c | c | c |}
    \hline
    \bf Method & \bf RMSE & \bf CC & \bf CCC \\ \hline
    CNN+RNN - W=100 - 1 layer  & 0.108 & 0.544 & 0.506  \\ 
    CNN+RNN - W=100 - 2 layers &  0.112 & 0.519 & 0.479 \\ 
    \bf CNN+RNN - W=100 - 3 layers & \bf 0.107 & \bf 0.554 & \bf 0.507 \\ \hline
    \end{tabular}
}
\end{center}
\label{tab:num_hidden_layers}
\end{table}

\subsection{Comparison with Other Techniques}
Table \ref{tab:other_methods} shows how our best performing CNN+RNN model compares to other techniques evaluated on the AV+EC 2015 dataset. Both our single frame CNN model with dropout and our CNN+RNN model achieve comparable or superior performance compared to the state-of-the-art techniques. Our single frame CNN model achieves a higher CCC value than the baseline\cite{ringeval2015av+} and is comprable with two other techniques \cite{chen2015multi, he2015multimodal}, all of which use temporal information. While our CNN+RNN model's performance is not quite as strong as the CNN+LSTM model of Chao et al. \cite{chao2015long}, in terms of CCC value, we would like to point out that the authors used a larger CNN on a larger external dataset. Specifically, the authors trained an AlexNet\cite{krizhevsky2012imagenet} on 110,000 images from 1032 people in the Celebrity Faces in the Wild (CFW) \cite{zhang2012finding} and FaceScrub datasets \cite{ng2014data}.


\begin{table}[t!]
\caption{Performance Comparison of Our Models versus Other Methods (D: Dropout, W: temporal window length)}
\begin{center}
\resizebox{0.95\columnwidth}{!}{
    \begin{tabular}{ | l | c | c | c |}
    \hline
    \bf Method & \bf RMSE & \bf CC & \bf CCC \\ \hline
	Baseline \cite{ringeval2015av+} & 0.117 & 0.358 & 0.273 \\ 
	LGBP-TOP + LSTM \cite{chen2015multi} & 0.114 & 0.430 & 0.354 \\ 
	LGBP-TOP+ Deep Bi-Dir. LSTM \cite{he2015multimodal} & 0.105 & 0.501 & 0.346 \\ 
	LGBP-TOP+LSTM+$\epsilon$-loss \cite{chao2015long} & 0.121 & 0.488 & 0.463 \\ 
	\bf CNN+LSTM+$\epsilon$-loss \cite{chao2015long} &\bf 0.116 & \bf 0.561 & \bf 0.538 \\ \hline
	\hline
    	Single Frame CNN+D - ours & 0.113 & 0.426 & 0.326 \\ 
       CNN+RNN - W=100 - 3 layers  - ours&  0.107 & 0.554 & 0.507 \\ \hline 
    \end{tabular}
}
\end{center}
\label{tab:other_methods}
\end{table}

\section{Conclusions}
\label{sec:conclusions}
In this work, we presented two systems for doing dimensional emotion recognition: a single frame CNN model and a multi-frame CNN+RNN model. We showed that our simple learned representation (single frame CNN) can outperform the baseline temporal model trained on hand-crafted features. With the CNN+RNN model, we showed how incorporating temporal information can yield smoother and more accurate predictions. Lastly, we conducted an extensive hyperparameter analysis and selected a CNN+RNN model that achieved comparable or superior performance to other state-of-the-art emotion recognition techniques on the AV+EC 2015 dataset.



\bibliographystyle{IEEEbib}
\bibliography{strings,refs}

\end{document}